\documentclass[runningheads,a4paper]{llncs}
\usepackage[utf8]{inputenc}
\usepackage[OT2,T1]{fontenc}

\usepackage[english,polish,russian]{babel}

\usepackage{todonotes}
\usepackage{times}
\usepackage{url}

\usepackage{graphicx,float,import,hyperref,placeins}

\usepackage{microtype} 

\usepackage{color}

\usepackage{amsmath,amssymb} 
\usepackage{wrapfig}
\usepackage{transparent}
\usepackage[normalem]{ulem}

\newcommand{\keywords}[1]{\par\addvspace\baselineskip
\noindent\keywordname\enspace\ignorespaces#1}

\newcommand{\R}[1]{\foreignlanguage{russian}{#1}}

\urldef{\mailsa}\path|mzapotoczny@gmail.com|
\urldef{\mailsb}\path|{pawel.rychlikowski,jan.chorowski}@cs.uni.wroc.pl|

\newsavebox{\foobox}
\newcommand{\slantbox}[2][.5]
  {%
    \mbox
      {%
        \sbox{\foobox}{#2}%
        \hskip\wd\foobox
        \pdfsave
        \pdfsetmatrix{1 0 #1 1}%
        \llap{\usebox{\foobox}}%
        \pdfrestore
      }%
  }

\begin{document}
\selectlanguage{english}
\title{On Multilingual Training of Neural Dependency Parsers}

\titlerunning{On Multilingual Training of Neural Dependency Parsers}

\author{Michał Zapotoczny \and Paweł Rychlikowski \and Jan Chorowski}


\authorrunning{Michal Zapotoczny et al.}

\institute{
Institute of Computer Science,
University of Wrocław,
Poland\\
\mailsa\\ \mailsb\\
}

\index{Zapotoczny, Michał}
\index{Rychlikowski, Paweł}
\index{Chorowski, Jan}

\toctitle{} \tocauthor{}

\maketitle

%
%
%
%
\begin{abstract}
  We show that a recently proposed neural dependency parser can be
  improved by joint training on multiple languages from the same
  family. The parser is implemented as a deep neural network whose
  only input is orthographic representations of words. In
  order to successfully parse, the network has to discover how linguistically
  relevant concepts can be inferred from word spellings. We analyze
  the representations of characters and words that are learned by the network to establish
  which properties of languages were accounted for. In
  particular we show that the parser has approximately learned to
  associate Latin characters with their Cyrillic counterparts and that
  it can group Polish and Russian words that have a similar
  grammatical function. Finally, we evaluate the parser on selected
  languages from the Universal Dependencies dataset and show that it
  is competitive with other recently proposed state-of-the art
  methods, while having a simple structure.

\end{abstract}

\section{Introduction}

Parsing text is an important part of many natural language processing
applications. Recent state-of-the-art results were obtained with
parsers implemented using deep neural networks
\cite{andor_globally_2016}. Neural networks are flexible learners
able to express complicated input-output relationships.
However, as more powerful machine learning techniques are used, the
quality of results will not be limited by the capacity of the
model, but by the amount of the available training data. In this
contribution we examine the possibility of increasing the training set
by using treebanks from similar languages.

For example, in the upcoming Universal Dependencies (UD) 2.0 treebank
collection~\cite{nivre_universal_2015} there are 863 annotated
Ukrainian sentences, 333 Belarusian, but nearly 60k Russian
ones (divided into two sets: a default one of 4.4k sentences and
SynTagRus with 55.4k sentences). Similarly, there are 7k Polish sentences and a little over 100k
Czech ones\footnote{However, experiments use UD 1.3 dataset
  which does not include Belarusian and Ukrainian.}.  Since these
languages belong to the same Slavic language family, performance on
the low resource languages should improve by joint training the model
also on a better annotated language \cite{bender_achieving_2011}.  In
this paper, we demonstrate this improvement. Starting with a parser
competitive with the current state-of-the-art, we are able to further
improve the results for tested languages from the Slavic family.  We
train the model on pairs of languages through simple parameter sharing
in an end-to-end fashion, retaining the structure and qualities of the
base model.

\section{Background and Related Work}
Dependency parsers represent sentences as trees in which every word
is connected to its head with a directed edge (called a
dependency) labeled with the dependency's type. Parsers often contain
parts that are learned on a corpus. In example, transition-based
dependency parsers use the learned component to guide their
actions, while graph-based dependency parser learn a scoring that
measures the quality of inserting a \emph{(head, dependency)} edge into the tree.

Historically, the learning algorithms were relatively simple
ones, e.g. transition-based parsers used linear SVMs 
\cite{nivre_maltparser_2005,nivre_algorithms_2008}. Recently, those 
simple learning models were successfully replaced by deep neural
networks~\cite{titov_latent_2007,chen_fast_2014,dyer_transition-based_2015,andor_globally_2016}. This
trend coincides with successes of those models on other
NLP tasks, such as language modeling
\cite{mikolov_recurrent_2010,jozefowicz_exploring_2016} and
translation
\cite{bahdanau_neural_2014,sutskever_sequence_2014,wu_google_2016}.

Neural networks have enough capacity to directly solve the
parsing task. For example a constituency parser can be implemented using a
sequence-to-sequence network originally developed for translation
\cite{vinyals_grammar_2014}. Similarly, a graph-based dependency parser can be
implemented by solving two supervised tasks: head selection and dependency
labeling. Both are easily solved using neural networks
\cite{kiperwasser_simple_2016,zhang_dependency_2016,dozat_deep_2016,chorowski_read_2016}.
Moreover, neural networks can extract meaningful features from the data,
which may augment or replace manually designed ones,
as it is the case with word embeddings~\cite{mikolov2013distributed}
or features derived from the spelling of words
\cite{kim_character-aware_2015,ballesteros_improved_2015,chorowski_read_2016}.

Another particularly nice property of neural models is that all
internal computations use distributed representations of input data
that are embedded in highly dimensional vector spaces
\cite{hinton_distributed_1986}. These internal representation can be
easily shared between tasks~\cite{caruana_multitask_learning}.
Likewise, neural parsers can share some of their parameters to harness
similarities between languages
\cite{bender_achieving_2011,guo2015cross,duong2015neural,ammar_many_2016}. Creation
of multilingual parsers is further facilitated by 
the introduction of standardized treebanks, such as the Universal
Dependencies \cite{nivre_universal_2015}.


\section{Model}\label{sec:model}

Our multilingual parser can be seen as $n$ identical neural
dependency parsers for $n$ languages, which share parameters. When all
parameters are shared a single parser is obtained for all $n$
languages. When only a subset of parameters is shared the model can be
seen as a parser for a main language that is partially regularized using data for
other languages.

\begin{figure}[t]
  \centering
  \resizebox{\textwidth}{!}{
    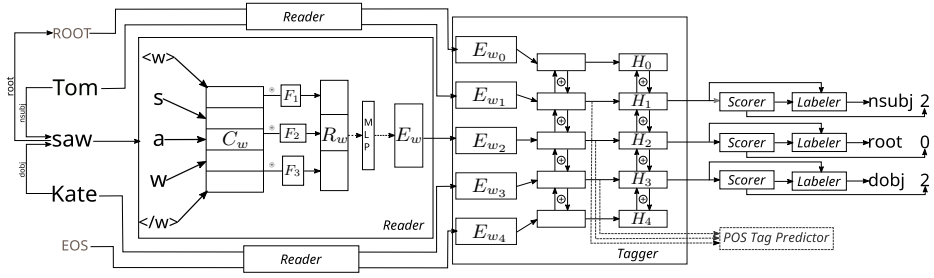
  }
  \caption{The model architecture.} 
  \label{fig:architecture}
\end{figure}

Each of the $n$ parsers is a single neural network
that directly reads a sequence of characters and finds dependency
edges along with their labels~\cite{chorowski_read_2016}. We can
functionally describe four basic parts: \emph{Reader},
\emph{Tagger}, \emph{Labeler/Scorer}, and an optional \emph{POS
  Tag Predictor} (Figure~\ref{fig:architecture}).

The \textbf{reader} is tasked with transforming the orthographic representation
of a single word $w$ into a vector $E_w \in \mathbb{R}^{\text{Edim}}$,
also called the word $w$'s embedding. 
%
First, we represent each word as a sequence of
characters fenced with start-of-word and end-of-word tokens. We
find low dimensional characters embeddings and concatenate them to
form a matrix $C_w$. Next we convolve this matrix
with a learned filterbank $F$
\begin{equation}
R_{w,i} = \max(C_w \ast F_i),
\end{equation}
where $F_i$ is the i-th filter and $\ast$ denotes convolution over the length
of the word. Thanks to the start- and end-of-word tokens the filters
can selectively target infixes, prefixes and suffixes of words. Finally, we max-pool the
filter activations over the word length and apply a
small feedforward network to obtain final word
embedding $E_w = \text{MLP}(R_w)$. 

The \textbf{tagger} processes complete sentences and puts individual word embeddings $E_w$ into
their contexts.
We use a multi-layer bidirectional GRU Recurrent Neural
Network (BiRNN) \cite{RNN_GRU,schuster_bidirectional_1997}. 
The output of the tagger is a sequence of the BiRNN's hidden states  $H_0, H_1, \dots, H_n$ with $H_i \in
\mathbb{R}^{\text{Hdim}}$, where $H_0$ corresponds to a prepended ROOT
word and
$n$ is the length of the sentence. Please observe that while the
embedding $E_i$ of the $i$-th word only depends on the word's spelling, the
corresponding hidden state $H_i$ depends on the whole sentence.

We have also added an \textbf{auxiliary network to predict POS tags} based
on hidden states $H_i$. It serves two purposes: first, it can provide
extra supervision on POS tags known during training. Second, it
helps to attribute errors to various parts of the network
(c.f. Sec. \ref{sec:error_analysis}). The POS tag
predictor is optional: its output is not used during inference because
the tagger communicates all information to the scorer and labeler
through the hidden states $H_i$.

Finally, the network produces the dependency tree by solving
two supervised learning tasks: using a \textbf{scorer} to find the
head word, then
using a \textbf{labeler} to find the edge label .

The \textbf{scorer} determines whether each pair of hidden vectors
$(H_w, H_h)$
forms a dependency. We employ per-word normalization of scores: for a given word location
$w\in {1,2,\dots,n}$ scores are SoftMax-normalzied over all head locations
$h \in {0,1,2,\dots,n}$.

The \textbf{labeler} reads a pair of hidden vectors $(H_w, H_h)$ and
predicts the label of this dependency edge.  During training we use
the ground-truth head location, while during inference we use the
location predicted using the \emph{scorer}.

We employ the following \textbf{training criterion}:
$$ L = \alpha_h L_h + \alpha_l L_l + \alpha_t L_t $$,
where $L_h$, $L_l$, $L_t$ are negative log-likelihood losses of
the scorer, the labeler and POS tag predictor, respectively. 

\section{Experiment Details and Results} \label{sec:results} 

\subsection{Model Hyperparameters}
We have decided to use the same set of hyperparameters for all
languages and multilingual parsers, which were a
compromise in model capacity for languages that had small and large
treebanks. The reported size of recurrent layers is slightly too
big for low-resources single-language parser, but we have determined
that it is optimal for languages with large treebanks and for
multilingual training. 

The \emph{reader} embeds each character into vector of size 15, and
contains 1050 filters (50$\cdot$k filters of length k for k = 1, 2,\dots, 6) 
whose outputs are projected into 512-dimensional vector transformed by a 3 equally
sized layers of feedforward neural network with ReLU activation.
Unlike \cite{kim_character-aware_2015,chorowski_read_2016} we decided
to remove Highway layers \cite{srivastava_highway_2015} from the
\emph{reader}. Their usage introduced a marginal accuracy gain, while
nearly doubling the computational burden.
The \emph{tagger} contains 2 BiRNN layers of GRU units with 548 hidden
states for both forward and backward passes which are later aggregated using
addition \cite{chorowski_read_2016}. Therefore the hidden states of the tagger are also 548-dimensional.
The \emph{POS tag predictor} consists of a single affine transformation
followed by a SoftMax predictor for each POS
category.
The \emph{scorer} uses a single layer of 384 tanh for head word
scoring while the \emph{labeller} uses 256 Maxout units
(each using 2 pieces) to classify the relation label
\cite{goodfellow_maxout_2013}. The training cost used the constants
$\alpha_h=0.6, \alpha_l=0.4, \alpha_t=1.0$.

We regularize the models using Dropout \cite{srivastava_dropout:_2014}
applied to the \emph{reader} output (20\%),  between the BiRNN
layers of the \emph{tagger} (70\%) and to the \emph{labeller} (50\%).
Moreover we apply mild weight decay of $0.95$.

We have trained all models using the Adadelta
\cite{zeiler_adadelta_2012} learning rule with epsilon annealed from
1e-8 to 1e-12 and adaptive gradient clipping
\cite{chorowski_end--end_2014}.  
Experiments are early-stopped on
validation set Unlabeled Attachment Score (UAS) score.  Unfortunately,
due to limited computational resources we are only able to present the
results for a subset of the UD treebanks that are shown in Table
\ref{tab:universal}.

Multilingual models use the same architecture. We unify the inputs and
outputs of all models by taking the union of all possible token
categories (characters, POS categories, dependency labels). If some
category does not exist within a particular language we use a special
UNK token. All parsers are trained in parallel minimizing a sum of
their individual training costs.  We use early-stopping on the main (first) language UAS score. We equalize training mini-batches such that each contains the same number of sentences from all languages. We determined the optimal amount of parameter sharing and show it in Table~\ref{tab:multi_baseline}. Moreover, we never share the start-of-word and end-of-word tokens to indicate to the network which language is parsed.  
\begin{table}[tb]
  \centering
  \caption{Baseline results of single language models from
    UD v1.3. Our models use only orthographic representations of
    tokenized words during inference and work without a separate POS tagger.
    Ammar et al. \cite{ammar_many_2016} uses version 1.2 of UD and uses gold
    language ids and predicted coarse tags.
    SyntaxNet\cite{andor_globally_2016,alberti_parsey_saurus_2017} works
    on predicted POS tags, while ParseySaurus\cite{alberti_parsey_saurus_2017}
    uses word spellings.}
  \label{tab:universal}
  \begin{tabular}{l l | l l | l l | c | l l}
    language & \#sentences & \multicolumn{2}{c|}{Ours} &
      \multicolumn{2}{c|}{SyntaxNet} & Ammar et al. & \multicolumn{2}{c}{ParseySaurus} \\ \hline
    & & UAS & LAS & UAS & LAS & LAS & UAS & LAS\\ \hline
    Czech & 87 913 & \textbf{91.41} & \textbf{88.18} & 89.47 & 85.93 & - & 89.09 & 84.99 \\
    Polish & 8 227 & 90.26 & 85.32 & 88.30 & 82.71 & - & \textbf{91.86} & \textbf{87.49}\\
    Russian & 5 030 & 83.29 & 79.22 & 81.75 & 77.71 & - & \textbf{84.27} & \textbf{80.65} \\
    German & 15 892 & 82.67 & 76.51 & 79.73 & 74.07 & 71.2 & \textbf{84.12} & \textbf{79.05}\\
    English & 16 622 & 87.44 & 83.94 & 84.79 & 80.38 & 79.9 & \textbf{87.86} & \textbf{84.45}\\ 
    French & 16 448 & \textbf{87.25} & \textbf{83.50} & 84.68 & 81.05 & 78.5 & 86.61 & 83.1\\
    Ancient Greek & 25 251 & \textbf{78.96} & \textbf{72.36} & 68.98 & 62.07 & - & 73.85 & 68.1
  \end{tabular}
\end{table}
  
\begin{table}[tb]
  \centering  
  \caption{Impact of parameter sharing strategies on main language parsing accuracy when multilingual training
    is used for additional supervision.}
  \label{tab:multi_baseline}
  \begin{tabular}{l | l l c c}
    Shared parts & Main lang & Auxiliary lang & UAS & LAS \\ \hline
      - & Polish & - & 90.26 & 85.32 \\
    \emph{Parser} & Polish & Czech & 90.72 & 85.57 \\ 
    \emph{Tagger, Parser} & Polish & Czech & 91.19 & 86.37 \\ 
    \emph{Tagger, POS Predictor, Parser} & Polish & Czech & 91.65 & 86.88 \\ 
    \emph{Reader, Tagger, POS Predictor, Parser} & Polish & Czech & \textbf{91.91} & \textbf{87.77} \\\hline 
    \emph{Parser} & Polish & Russian & 90.31 & 85.07 \\  
    \emph{Tagger, POS Predictor, Parser} & Polish & Russian &
    \textbf{91.34} & \textbf{86.36} \\ 
    \emph{Reader, Tagger, POS Predictor, Parser} & Polish & Russian & 89.16 & 82.94 \\  
\hline\hline 
    - & Russian & - & 83.29 & 79.22\\
    \emph{Parser} & Russian & Czech & 83.15 & 78.69 \\  
    \emph{Tagger, POS Predictor, Parser} & Russian & Czech & 83.91 & 79.79 \\ 
    \emph{Reader, Tagger, POS Predictor, Parser} & Russian & Czech & \textbf{84.78} & \textbf{80.35} 
  \end{tabular}
\end{table}

\subsection{Main Results}

Our results on single language training are presented in
Table~\ref{tab:universal}. Our models reach better scores than the
highly tuned SyntaxNet transition-based parser
\cite{andor_globally_2016} and are competitive with the DRAGNN based
ParseySaurus which also uses character-based input
\cite{alberti_parsey_saurus_2017}.

Multilingual training
(Table~\ref{tab:multi_baseline}) improves the performance on
low-resource languages. We observe that the optimal amount of
parameter sharing depends on the similarity between languages and
corpus size -- while
it is beneficial to share all parameters of the PL-CZ and RU-CZ parser, the
PL-RU parser works best if the reader subnetworks are separated. We attribute this
to the quality of Czech treebank which has several times more examples than Polish and
Russian datasets combined.

\subsection{Analysis of Language Similarities Identified by the
  Network}
We have first analyzed whether a PL-RU parser can learn the
correspondence between Latin and Cyrillic
scripts\footnote{Conveniently, the Unicode has separate codes for
  Latin and Cyrillic letters.}. We have inspected the reader
subnetworks of a PL-RU parser that shared all parameters. As described
in Section \ref{sec:model}, the model begins processing a word by finding
the embedding of each character. For the analysis we have extracted
the embeddings associated with all Polish and Russian characters.
We have paired Polish and Russian letters
which have similar pronunciations. We note that the pairing omits
letters that have no clear counterparts (e.g. the Russian letter \R{я}
correspond to the syllable ``ja'' in Polish). 
\begin{quote}
  a-\R{а}, b-\R{б}, c-\R{ц}, d-\R{д}, e-\R{е}, e-\R{э}, f-\R{ф},
  g-\R{г}, h-\R{х}, i-\R{и}, j-\R{й}, k-\R{к}, l-\R{л}, m-\R{м},
  n-\R{н}, o-\R{о}, p-\R{п}, r-\R{р}, s-\R{с}, t-\R{т}, u-\R{у},
  w-\R{в}, y-\R{ы}, z-\R{з}, ł-\R{л}, ż-\R{ж}
\end{quote}

Adapting the famous equation $\text{\emph{king}}-\text{\emph{man}} + \text{\emph{woman}}
\approx \text{\emph{queen}}$ \cite{mikolov2013distributed} we inspected to what
extent our network was able to deduce Latin-Cyrillic
correspondences. For all distinct pairs $(p_1-r_1, p_2-r_2)$ of letter
correspondences we computed the vector $C(p_2) - C(p_1) + C(r_1)$,
where $C$ stands for char embedding, and found Russian letter which had
the closest embedding vector. In 48.3\% cases we choose the right
vector. We found it quite striking given that the two languages have
separated from their common root (Proto-Slavic) more than 1000 years
ago. Moreover, relations between Polish and Russian letters are side
effects, not the main objective of the neural network.


We have also examined word representations $E_w$ computed for Polish
and Russian by the shared reader subnetwork. As one could expect, the
network was able to realize that in these languages morphology is
suffix based.
%
%
%
However, the network was also able to learn that words built from
different letters can behave in similar way.
We can observe it in both monolingual or multilingual context. Table
\ref{tab:pol-ru-words} shows some Polish adjectives and the top-7
Russian words with the closest embedding. All Russian words which are
not \emph{italics} have the same morphological tags as the Polish 
word. In the first row we can observe 2 suffixes \slantbox[.25]{\R{-ской}}
(skoy) and \slantbox[.25]{\R{-нной}} (nnoy) quite distant from polish
\emph{-owej} (ovey). In the second row we see that the model was able
to correctly alias the Polish 3-letter suffix \emph{-ych} with the
Russian 2 letter suffix \slantbox[.25]{\R{-ых}} which are pronounced the same
way. The relation found by the network is purely syntactical -- there
is no easy-to-find connection between semantics of these words.

\begin{table}[tb]
\centering
\caption{The network learns to group Polish words with Russian words
  that have a similar grammatical function.}\label{tab:pol-ru-words}
\begin{tabular}[pos]{l|l}
  Polish word & Closest Russian embeddings \\ \hline
  przedwrześniowej & \R{адренергической тренерской таврической} \\ 
                   & \R{непосредственной археологической философской \slantbox[.25]{верхнюю}} \\ \hline
  większych        & \R{автомобильных \slantbox[.25]{трёхдневные} технических} \\
                   & \R{практических официальных оригинальных} \\ \hline
  policyjnym       & \R{главным историческим глазным} \\
                   & \R{непосредственным \slantbox[.25]{косыми} летним двухсимвольным} \\
\end{tabular}
\end{table}

\subsection{Common Error Analysis}\label{sec:error_analysis}


We have investigated two possible sources of errors produced by the
parser. First, we verified if using a more advanced tree-building
algorithm was better than using a greedy one. We have observed that the \emph{scorer} produces very sharp probability distributions that
can be transformed into trees using a greedy algorithm that
simply selects for each word the highest scored head
\cite{chorowski_read_2016,dozat_deep_2016}. Counterintuitively, the Chu-Liu-Edmonds (CLE)
maximum spanning tree algorithm \cite{edmonds_optimim_1966} often makes the
decoding results slightly worse. We have established that the network
is so confident in its predictions that non-top
scores do not reflect alternatives but
are only noise. Therefore when the greedy decoding creates a
cycle the CLE  usually breaks it in a wrong place introducing another
pointer error.

We have used the \emph{POS predictor} to
pinpoint which parts of the network (\emph{reader/tagger} or \emph{labeler/scorer})
were responsible for errors. Tests showed
that if the predicted tag was wrong, the
\emph{scorer} and \emph{labeler} will nearly always  produce erroneous results too. 

\section{Conclusions and Future Works}
We have demonstrated a graph-based dependency parser implemented as a
single deep neural network that directly produces parse trees from
characters and does not require other NLP tools such as a POS
tagger. The proposed parser can be easily used in a multilingual
setup, in which parsers for many languages that share parameters are
jointly trained. We have established that the degree of sharing
depends on language similarity and corpus size: the best PL-CZ parser
and RU-CZ shared all parameters (essentially creating a single parser for both
languages), while the best PL-RU parser had
separate morphological feature detectors (i.e. \emph{readers}).  We
have also determined that the network can extract meaningful relations
between languages, such as approximately learning a mapping from Latin to
Cyrillic characters or associate Polish and Russian words that have a
similar grammatical function. While this contribution focused on
improving the performance on a low-resource language using data from
another languages, similar parameter sharing techniques could be used
to create one universal parser \cite{ammar_many_2016}.

We have performed qualitative error analysis and have determined to
regions for possible future improvements. First, the network does not
indicate alternatives to the produced parse tree. Second, errors in
word interpretation are often impossible to correct by the upper
layers of the network. In the future we plan to investigate training a
better POS tagging subnetwork possibly using other sources of data.

\textbf{Acknowledgments.}
The experiments used Theano
\cite{bergstra+al:2010-scipy}, 
Blocks and Fuel \cite{vanmerrienboer_blocks_2015} libraries.
The authors would like to acknowledge the support of the following
agencies for research funding and computing support: National Science
Center (Poland) grant Sonata 8 2014/15/D/ST6/04402, National Center
for Research and Development (Poland) grant Audioscope (Applied Research Program,
3rd contest, submission no. 245755). 

\bibliographystyle{splncs03}
\bibliography{parser}

\end{document}